\documentclass[letterpaper, 10 pt, conference]{ieeeconf}
\IEEEoverridecommandlockouts
\usepackage{cite}
\usepackage{float}
\usepackage{booktabs}
\usepackage{multirow}
\usepackage{amsmath,amssymb,amsfonts}
\usepackage{algorithmic}
\usepackage{graphicx}
\usepackage{textcomp}
\usepackage{soul} 
\usepackage{makecell}
\usepackage{xcolor}

\begin{document}

\title{\LARGE \bf UAVD-Mamba: Deformable Token Fusion Vision Mamba for Multimodal UAV Detection
\author{Wei Li$^{1}$, Jiaman Tang$^{1}, $Yang Li$^{1}$, Beihao Xia$^{2}$, Ligang Tan$^{1}$, Hongmao Qin$^{1}$}

\thanks{This work is supported by the National Key Research and Development Program of China under Grant 2023YFB2504701, 2023YFB2504704, and the Open Project Program of Fujian Key Laboratory of Special Intelligent Equipment Measurement and Control under Grant No.FJIES2024KF07.  (\textit{Corresponding Author: Yang Li, Ligang Tan})}
\thanks{$^{1}$ Wei Li, Jiaman Tang, Yang Li, Ligang Tan, Hongmao Qin are with the College of Mechanical and Vehicle Engineering, Hunan University, Changsha 410082, China. (email: great\_plum@163.com; tjm86464@gmail.com; lyxc56@gmail.com; tlg9@163.com; 
qinhongmao@vip.sina.com)}
\thanks{$^{2}$ Beihao Xia, School of Electronic Information and Communications, Huazhong University of Science and Technology, Wuhan, Hubei 430074, China, and also with Fujian Key Laboratory of Special Intelligent Equipment Safety Measurement and Control, Fujian Special Equipment Inspection and Research Institute, Fuzhou 350008, China. (email: xbh\_hust@hust.edu.cn)
}
}


\maketitle
\thispagestyle{empty}
\pagestyle{empty}
\begin{abstract}
Unmanned Aerial Vehicle (UAV) object detection has been widely used in traffic management, agriculture, emergency rescue, etc. However, it faces significant challenges, including occlusions, small object sizes, and irregular shapes. These challenges highlight the necessity for a robust and efficient multimodal UAV object detection method. Mamba has demonstrated considerable potential in multimodal image fusion. Leveraging this, we propose UAVD-Mamba, a multimodal UAV object detection framework based on Mamba architectures. 
To improve geometric adaptability, we propose the Deformable Token Mamba Block (DTMB) to generate deformable tokens by incorporating adaptive patches from deformable convolutions alongside normal patches from normal convolutions, which serve as the inputs to the Mamba Block. To optimize the multimodal feature complementarity, we design two separate DTMBs for the RGB and infrared (IR) modalities, with the outputs from both DTMBs integrated into the Mamba Block for feature extraction and into the Fusion Mamba Block for feature fusion. Additionally, to improve multiscale object detection, especially for small objects, we stack four DTMBs at different scales to produce multiscale feature representations, which are then sent to the Detection Neck for Mamba (DNM). The DNM module, inspired by the YOLO series, includes modifications to the SPPF and C3K2 of YOLOv11 to better handle the multiscale features. In particular, we employ cross-enhanced spatial attention before the DTMB and cross-channel attention after the Fusion Mamba Block to extract more discriminative features. Experimental results on the DroneVehicle dataset show that our method outperforms the baseline OAFA method by 3.6\% in the mAP metric. Codes will be released at https://github.com/GreatPlum-hnu/UAVD-Mamba.git.





\end{abstract}



\section{Introduction}
UAV object detection has received wide attention in traffic management and urban governance\cite{YUAN2024102246}. However, it faces several challenges, such as occlusion by trees or buildings, small target sizes, irregular shapes, shadows, etc. Traditional methods based on a single modality often struggle with low accuracy, limited generalization, and high sensitivity to noise. Multimodal approaches also face issues like multimodal misalignment, data redundancy, and suboptimal integration of complementary information. These challenges highlight the need for a more accurate, efficient, and robust multimodal UAV object detection method.

\begin{figure}
    \centering
    \setlength{\abovecaptionskip}{-0.2cm}
    \includegraphics[width=1\linewidth]{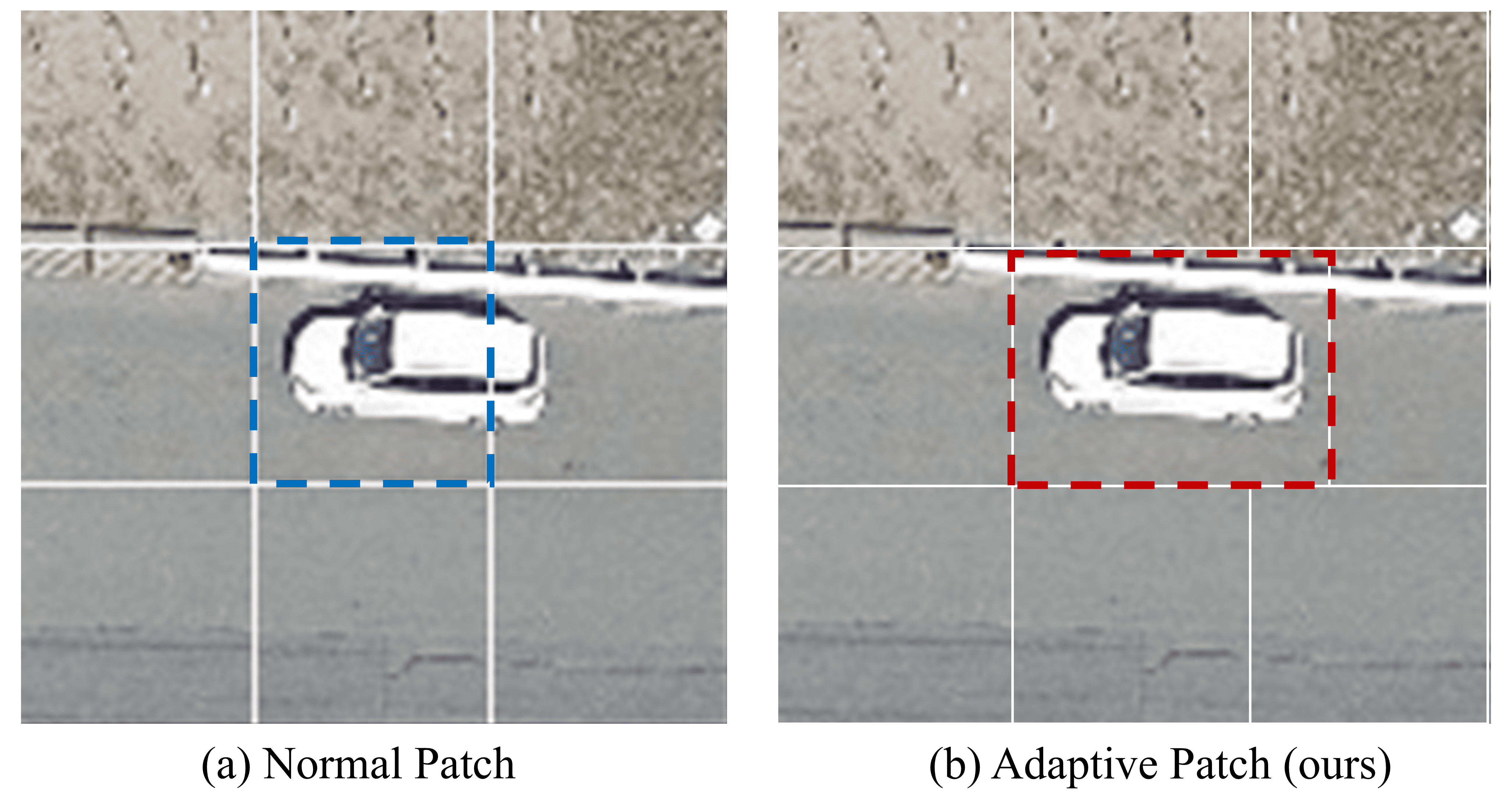}
    \caption{Previous Vision Mamba\cite{gu2023mamba} split the input image into the normal patch, and our UAVD-Mamba split into the adaptive patch. (a) Normal patch (blue dashed rectangular box), which uses convolution kernels with a stride equal to the patch size to split the input image into patches. (b) Adaptive patch (red dashed rectangular box), which uses deformable convolutions to split the input image into patches that can enhance geometric adaptability and obtain more discriminative features.}
    \label{fig:token}
\end{figure}

\begin{figure*}[t!]
\vspace{5pt} 
\centering
\includegraphics[width=1\textwidth]{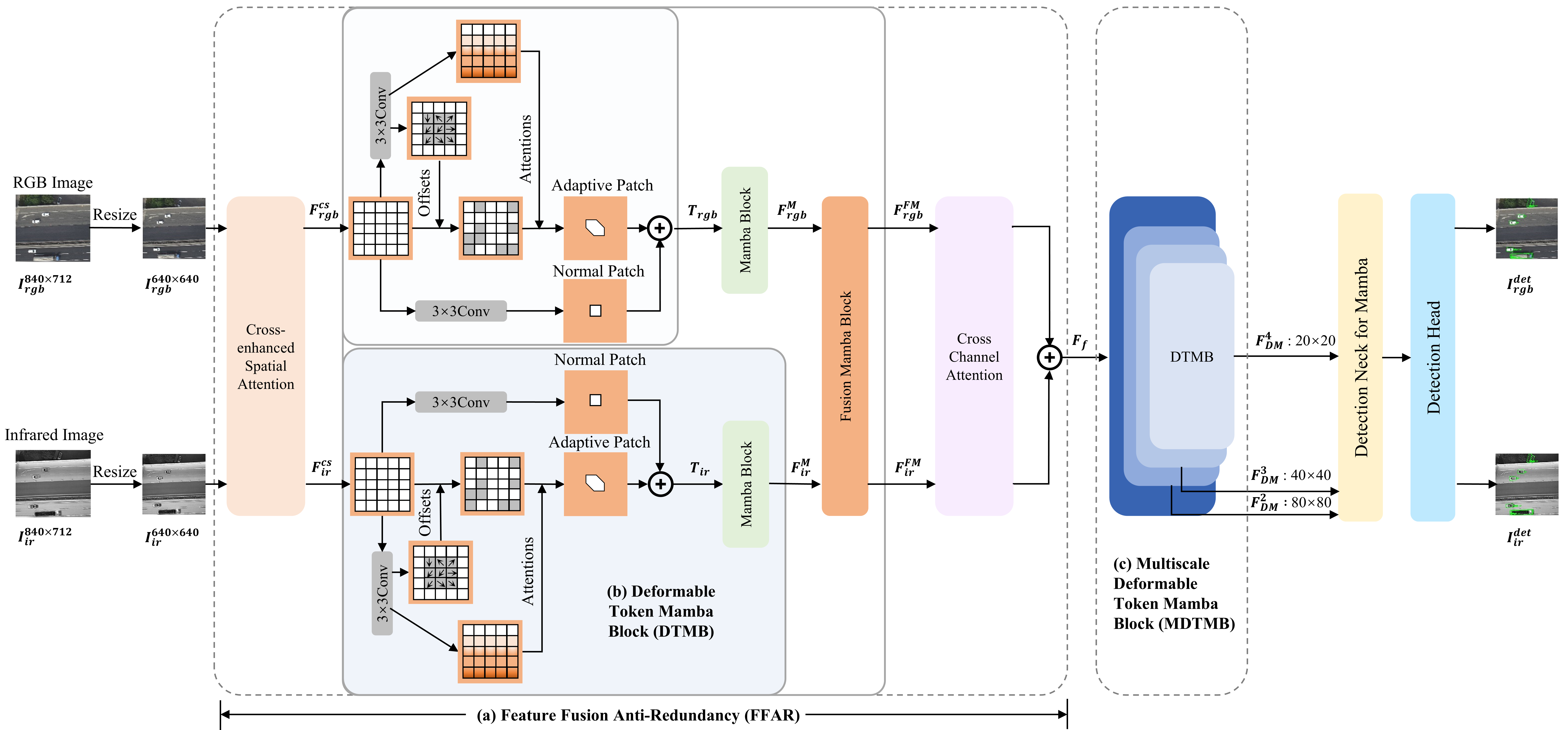}
\caption{An Overview of UAVD-Mamba. The RGB-IR image pairs are first resized and then sent to the FFAR for multimodal feature fusion. FFAR consists of four modules, including Cross-enhanced Spatial Attention, Deformable Token Mamba Block (DTMB), Fusion Mamba Block, and Cross Channel Attention. In FFAR, after being enhanced by spatial attention, these features are passed to the DTMB module and Fusion Mamba Block for multimodal fusion feature extraction. In particular, we design two separate DTMBs for RGB and IR modalities to improve the multimodal feature complementarity. In each DTMB, we use a deformable convolutional layer to generate the adaptive patch and the normal patch to form the deformable token, serving as inputs for the Mamba Block. The outputs of Mamba Blocks of the RGB and IR branches are integrated into the Fusion Mamba Block for feature fusion. Cross-channel attention further optimizes feature fusion and reduces redundancy.
For multiscale object detection, the Multiscale Deformable Token Mamba Block (MDTMB) is designed by stacking four DTMBs for multiscale feature extraction. Finally, the multiscale fusion features are then sent to the Detection Neck for Mamba, and the detection head to generate the final detection results. }
\label{fig:Method}
\end{figure*}

UAV object detection methods mainly use convolutional neural networks (CNNs)\cite{li2017scale,agrawal2020multi,zhang2018deconv,sai2019object,kayalibay2017cnn} and transformer-based models\cite{hu2022fusformer,peng2023u2net}. CNN-based methods exhibit limitations in handling long-range dependencies, and transformer-based models suffer from high computational complexity. 
Mamba\cite{zhu2024vision}, with its efficient modeling and balanced complexity, has shown exceptional performance in computer vision. In particular, Mamba also demonstrates its great potential in multimodal fusion\cite{li2024coupled,ren2024remotedet}, while significantly improving computational efficiency. The shapes of the targets are usually irregular, requiring the object detector to have geometric adaptability. However, when Mamba executes visual tasks\cite{cao2024novel,chen2403mim,wang2024mask}, it typically employs a fixed partition strategy and cannot adaptively adjust the patching strategy to adapt to irregularly shaped objects\cite{shen2024famba,zhou2025mamba}, causing a loss of information integrity for individual tokens and subsequently impacting the accuracy of feature representation. \par

To enhance the geometric adaptability for UAV detection, we propose UAVD-Mamba, a multimodal UAV object detection framework based on Mamba architectures. Specifically, we introduce the Deformable Token Mamba Block (DTMB), which uses deformable and normal convolutions to generate adaptive and normal patches. The normal patch and adaptive patch are shown in Fig.~\ref{fig:token}. These patches are then fused to construct deformable tokens for improving feature representation. To optimize performance for each modality, we design two separate DTMBs—one for RGB and one for infrared. For multiscale object detection, DTMBs are stacked at different scales and processed by the Detection Neck for Mamba (DNM), which incorporates YOLOv11-inspired modifications. Additionally, cross-enhanced spatial and channel attention further refine feature extraction, boosting accuracy and discrimination.
Our contributions are summarized as follows:
 \begin{itemize}
\item We propose UAVD-Mamba, a multimodal UAV
object detection framework based on Mamba architectures, leveraging the adaptive deformable token and the multiscale detection module for Mamba to improve accuracy and robustness while reducing data redundancy.
\item To enhance geometric adaptability, we generate deformable tokens by incorporating adaptive patches from deformable convolutions alongside normal patches from convolutions, which serve as the inputs to the Mamba Block. Two separate Deformable Token Mamba Blocks (DTMB) for RGB and infrared (IR) modalities are built to strengthen the multimodal feature complementarity.
\item To enable multiscale object detection, we stack four DTMBs at different scales and propose the Detection Neck for Mamba (DNM), incorporating specific modifications to the SPPF and C3K2 of YOLOv11 to better process features extracted by the Mamba modules.
\end{itemize} 

The study is organized as follows. Section II reviews related works. Section III gives an overview of the structure of our model and then describes the main modules. Section IV presents the experimental setup and results. Finally, the concluding remarks are given in Section V.




\section{Related Work}
\subsection{UAV Object Detection}
Many UAV object detection methods have been proposed over the years, including single-modal approaches and multi-modal approaches.
In single-modal approaches, \cite{hu2023improving} proposed an anchor box optimization method for small object detection, and \cite{zhang2023oriented} designed an infrared enhancement framework using a kaleidoscope module and semantic feature supplementation. 
For multi-modal approaches, \cite{chen2024drone} leveraged a Transformer backbone with visual prompts for RGB-IR feature extraction. \cite{ouyang2024multimodal} and \cite{yuan2024c} enhanced RGB-T/IR fusion via cross-modal interaction and cross-attention, while \cite{cheng2023slbaf} employed adaptive fusion for improved robustness. To address redundancy and modality gaps, \cite{wang2024multi} proposed a fusion feature optimization network, and \cite{chen2024weakly} introduced spatial offset modeling with deformable alignment for better RGB-IR matching. However, 
Feature-level multi-modal fusion might suffer from feature misalignment \cite{cheng2023slbaf} and data redundancy, while decision-level fusion\cite{kim2016robust} is affected by inconsistencies in model results.

\subsection{Mamba for Computer Vision Tasks}
Mamba has shown great potential in visual tasks such as multimodal fusion and small object detection \cite{zhu2024vision}. \cite{li2024coupled} designed the Cross-modal Fusion Mamba (CFM) module based on Mamba's SS2D mechanism, enhancing small object distinguishability and improving class discrimination using local information. \cite{ren2024remotedet} applied Coupled Mamba to multimodal fusion, significantly improving its efficiency and accuracy. \cite{chen2403mim} explored Mamba for infrared small target detection (ISTD), treating local patches as visual sentences to capture global information using the outer Mamba layer, thereby enhancing Mamba's ability to capture critical local features. However, as image data is represented as pixel matrices, which lack the inherent tokenization structure present in textual data\cite{gu2023mamba}, it's difficult to design appropriate tokens for Mamba with image processing. Current research utilizing Mamba\cite{cao2024novel,chen2403mim,wang2024mask} for feature extraction divides images into fixed square regions for tokenization, which reduces token integrity and feature accuracy, as well as ignores the irregularity of object shapes\cite{garcia2022quantum}.
\section{Method}
In this section, we provide an overview of the proposed method, UAVD-Mamba, and then introduce the main components in our framework.
\subsection{Overview}\label{AA}
Our approach seeks to improve the geometric adaptability and multimodal feature extraction ability by leveraging Mamba architectures in UAV object detection. As shown in Fig.~\ref{fig:Method}, a pair of RGB-IR images is fed into a dual-stream network, with the image size adjusted to a preset input dimension. To obtain bimodal complementary features, we design two separate Deformable Token Mamba Blocks (DTMB) for the
RGB and infrared (IR) modalities, where adaptive patches generated by deformable convolutions are added to normal patches to form deformable tokens that serve as inputs to the Mamba. To enable multiscale object detection, we stack four DTMBs at different scales and propose the Detection
Neck for Mamba (DNM), incorporating specific modifications to better
process features extracted by the DTMB. Spatial and channel attention mechanisms are applied both before and after the Fusion Mamba Block (FMB) to enhance feature integration and reduce redundancy. 

\subsection{Feature Fusion Anti-Redundancy Module}
As shown in Fig.~\ref{fig:Method}(a), we put the Cross-enhanced Spatial Attention, DTMB, Fusion Mamba Block, and the Cross Channel Attention together to build a module called Feature Fusion Anti-Redundancy (FFAR). 
This module aims to promote feature fusion complementarity while reducing data redundancy.

\textbf{Cross-enhanced Spatial Attention.}  The Cross-enhanced Spatial Attention sub-module locates key spatial regions in RGB and IR images by cross-analyzing their spatial features, enhancing attention allocation to critical regions, and improving the expression of image features. For each RGB-IR image pair, the image size is first resized to a square ($I_{rgb}\in \mathbb{R} \mathbb{} ^{H\times W\times 3}$ and $I_{ir}\in \mathbb{R} \mathbb{} ^{H\times W\times 1}$), and then $I_{rgb}$ and $I_{ir}$ are passed through the spatial attention mechanism to obtain the spatial attention $F _{m}^{s}$ for each modality:
\begin{equation}
F _{m}^{s}=\sigma \left ( f_{i\times i} \left ( Cat\left ( Max\left ( I_{m}  \right ), Mean\left ( I_{m}  \right ) \right )  \right )  \right )\label{eq}
\end{equation}
where $m\in \left \{ rgb,ir \right \} $, $\sigma$ is the sigmoid function, $f_{i\times i}$ denotes the $i\times i$ convolution layer, $Cat\left ( \cdot  \right )$ denotes the concatenation operation, $Max\left ( \cdot  \right )$ denotes the maximum value and $Mean\left ( \cdot  \right )$ denotes the average value along the channel dimension. 

Conventional RGB-IR multimodal attention mechanisms typically rely on mutually exclusive division formulas\cite{cheng2023slbaf}. However, the features of the two modalities should enhance each other. Therefore, we multiply image $I_{m}$, the RGB spatial attention $F _{rgb}^{s}$, and the IR spatial attention $F _{ir}^{s}$ to enhance feature extraction, and obtain the enhanced spatial attention $F_{m}^{cs}$ of each modality:
\begin{equation}
F_{m}^{cs}=I_{m}\otimes{F}_{rgb}^{s}\otimes{F}_{ir}^{s}\label{eq}
\end{equation}
where $\otimes$ denotes element-wise product operation.

\begin{figure*}[t!]
\vspace{5pt} 
\centering
\setlength{\abovecaptionskip}{-0.1cm}
\includegraphics[width=1\textwidth]{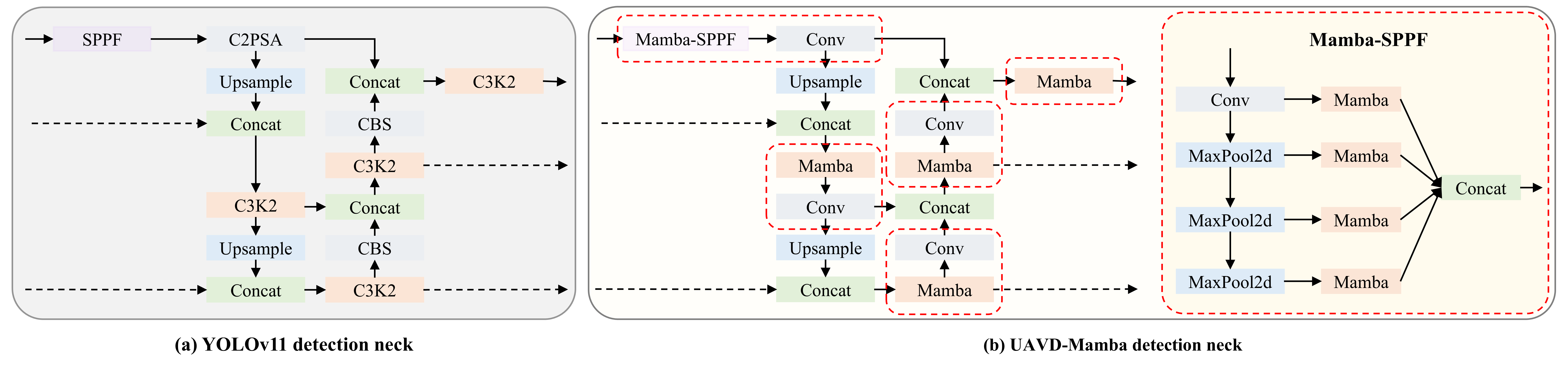}
\caption{We propose the Detection Neck for Mamba (right side), incorporating specific modifications to the SPPF and C3K2 of the neck of YOLOv11 (left side) to better process features extracted by the DTMB. The modified areas are highlighted with red dashed rectangles.}
\label{fig:detection neck}
\end{figure*}

\textbf{Deformable Token Mamba Block.}  Mamba Block uses a fixed division to partition the image. This fixed-size patch division can disrupt the integrity of individual token information, which negatively impacts the accuracy of feature representation. As shown in Fig.~\ref{fig:Method}b, we construct deformable tokens by integrating adaptive patches and normal patches. This operation can dynamically adjust the patch size according to the image content, generating patches of varying shapes and enhancing image feature extraction. During the computation process, the results of convolution $Conv\left ( \cdot  \right )$ and deformable convolution $DConv\left ( \cdot  \right )$\cite{zhu2019deformable} are added to efficiently control computational complexity and optimize the gradient backpropagation while guaranteeing effective feature extraction. The formulas are as follows:
\begin{equation}
T_{m}=Conv\left ( {F} _{m}^{cs}  \right )+DConv\left ( {F} _{m}^{cs}  \right )\label{eq}
\end{equation}
where $T_{m}$ denotes the result obtained after patching ${F} _{m}^{cs}$ through deformable tokens.

The deformable tokens are fed into the Mamba Block to obtain each modal feature ${F} _{m}^{M}$ that is preliminarily processed by the Mamba Block:
\begin{equation}
{F} _{m}^{M}=Mamba\left ( T_{m}  \right )\label{eq}
\end{equation}
where $Mamba\left ( \cdot  \right )$ denotes a Vision Mamba Block that is flattened using a four-way sequence modeling approach and combined with a residual network. Refer to \cite{PengSR2024} for more details.

\textbf{Fusion Mamba Block.}
To fully utilize the complementarity of the RGB and IR features, each modal feature is fed into the Fusion Mamba Block, and with the help of the state transfer equation provided by the other modal feature, the information of each modality feature is fully supplemented, and the complementary modal feature ${F} _{m}^{FM}$ is obtained:
\begin{equation}
{F} _{rgb}^{FM}=FusionMamba\left ( {F} _{rgb}^{M},{F} _{ir}^{M} \right )\label{eq}
\end{equation}
\begin{equation}
{F} _{ir}^{FM}=FusionMamba\left ({F} _{ir}^{M},{F} _{rgb}^{M} \right )\label{eq}
\end{equation}
where $FusionMamba\left ( \cdot  \right )$ denotes the Fusion Mamba Block, which has two inputs and can extend the original state space model (SSM) to a fusion FSSM. Compared to traditional single-input SSM, the former input in FSSM is the sequence to be processed, and the latter input generates the projection and time scale parameters.\par
\textbf{Cross Channel Attention.}  The complemented feature ${F} _{m}^{FM}$ is fed into the channel attention mechanism to obtain the channel attention ${F} _{m}^{c}$ for each modality, ${F} _{m}^{c}$ is denoted as:
\begin{equation}
{F} _{m}^{c} = \sigma \left ( f_{mlp} \left ( AvgPool\left ( F_{m}^{FM}  \right )  \right ) \right ) 
\end{equation}
where $f_{mlp}$ denotes the shared multilayer perceptron, and $AvgPool\left ( \cdot  \right )$ denote maximum pooling.

In traditional methods, channel attention for each modality is typically concatenated along the channel dimension. However, this approach often leads to significant information redundancy and fails to effectively capture complementary information between modalities. Therefore, we propose a cross-channel attention scheme. The following operation is performed for each modality. First, the complemented feature ${F} _{m}^{FM}$ is multiplied by its own channel attention, and then divided by the channel attention of the other modality. Finally, the results from both modalities are summed to obtain the cross-channel attention feature ${F} _{f}$, denoted as:
\begin{equation}
{F} _{f}= \frac{{F} _{rgb}^{FM}\times {F} _{rgb}^{c}}{{F} _{ir}^{c}}+\frac{{F} _{ir}^{FM}\times {F} _{ir}^{c}}{{F} _{rgb}^{c}}\label{eq}
\end{equation}
The cross-channel attention feature ${F} _{f}$ is sent to the following Multiscale Deformable Token Mamba Block module. ${F} _{f}$ is the output of FFAR.

\subsection{Multiscale Deformable Token Mamba Block Module}
We stack four DTMBs at different scales to enhance multiscale object detection by adjusting the step size of patching in DTMB, as shown in Fig.\ref{fig:Method}c. We put the feature ${F} _{f}$ into the DTMB and iterated it four times. The output of the first DTMB is used as the input for the next DTMB. The update process is written as:
\begin{equation}
F _{DM}^{1}=DTMB\left ({F} _{f} \right )\label{eq}
\end{equation}
\begin{equation}
F _{DM}^{n+1}=DTMB\left ({F} _{DM}^{n} \right )\label{eq}
\end{equation}
where n=1,2,3, ${F} _{DM}^{n}$ represents the features after the nth pass through the DTMB. Overall, in the study, we choose ${F} _{DM}^{2}$, ${F} _{DM}^{3}$ and ${F}_{DM}^{4}$ as inputs of Detection Neck for Mamba.\par
\subsection{Detection for Vision Mamba}
Our detection neck module is inspired by the YOLO series and incorporates specific modifications to adapt to the multiscale features extracted by DTMB. Specifically, the C3K2 module in the YOLO detection neck is replaced with the Mamba Block to fully utilize the advantages of the Mamba architecture, as shown in Fig.~\ref{fig:detection neck}. In addition, the original SPPF module applies the Mamba Block to the features at each scale after max pooling. These enhancements help improve detection performance. The features obtained from the Detection Neck for the Mamba Block are eventually passed into the Detection Head of YOLOv11. Our loss function is similar to YOLOv11\cite{yolov11}, the total loss function $L_{total}$ is composed of classification loss $L_{cls}$, box loss $L_{box}$, and distribution focal loss $L_{dfl}$: 
\begin{equation}
L_{total}=\lambda _{clc}L_{cls}+\lambda _{box}L_{box}+ \lambda _{dfl}L_{dfl}\label{eq}
\end{equation}
where $\lambda_{cls}$, $\lambda _{box}$, and $\lambda _{dfl}$ are the coefficients for each loss term.

\begin{table*}[!ht]
\vspace{10pt} 
\centering
\caption{Detection results (mAP, in \%) on DroneVehicle dataset. Note that all detectors locate and classify vehicles with OBB heads. The best results are highlighted in \textbf{bold}. And the second one is marked with \underline{underline}.} 
\setlength{\tabcolsep}{4.5mm}
\label{tab:tableTab} 
\begin{tabular}{cccccccc}
\toprule 
Detectors                                             & Input Category                                    & Car   & Truck & Freight-car & Bus   & Van   & mAP (\%)↑ \\ 
\midrule
YOLOv11 (Base) \textcolor{red}{(Github'24)}                                               & \multirow{2}{*}{RGB}                                             & 96.4  & 74.4  & 54.2        & 95.0    & 56.3  & 75.3       \\ 
Hu et al. \textcolor{red}{(RS'23)} \cite{hu2023improving}                                 &                         & 96.2 & 75.8 & 57.3       & 94.5 & 56.7 & 76.1       \\
\midrule
DAIK \textcolor{red}{(TRGS'23)} \cite{wang2023directional} & \multirow{4}{*}{IR}                                           & 90.2 & 71.6  & 57.4       & 89.9 & 50.2  & 71.7 \\
I$^2$MDet \textcolor{red}{(TRGS'23)} \cite{zhang2023oriented}                              &                                             & 96.3  & 73.4  & 65.0          & 93.2  & 58.6  & 77.3       \\
YOLOv11 (Base) \textcolor{red}{(Github'24)}                                               &                                             & \underline{98.3}  & 77.5  & 65.8        & 95.0    & 59.9  & 79.3       \\
Hu et al. \textcolor{red}{(RS'23)} \cite{hu2023improving}                                 &                          & 98.0 & \underline{79.5} & 67.2       & \underline{94.8} & 58.6 & \underline{79.6}      \\
\midrule
VIP-Det \textcolor{red}{(Drones'24)} \cite{chen2024drone}                                               & \multicolumn{1}{c}{\multirow{8}{*}{RGB+IR}}                                        & 90.4  & 78.5  & 61.4        & 89.8  & 57.5  & 75.5       \\ 
M2FP \textcolor{red}{(J-STARS'24)} \cite{ouyang2024multimodal}                                                  & \multicolumn{1}{c}{}                                     & 95.7  & 76.2  & 64.7        & 92.1  & 64.7  & 78.7       \\ 
C$^2$Former \textcolor{red}{(TGRS'24)} \cite{yuan2024c}                                             & \multicolumn{1}{c}{}                      & 90.2  & 68.3  & 64.4        & 89.8  & 58.5  & 74.2       \\
SLBAF \textcolor{red}{(MTA'24)} \cite{cheng2023slbaf}                                                 & \multicolumn{1}{c}{}  & 97.4  & 75.4  & 62.6        & 94.8  & 52.6  & 76.6       \\
Wang et al. \textcolor{red}{(J-STARS'24)} \cite{wang2024multi}                                              & \multicolumn{1}{c}{}                        & 90.4  & 72.6  & 68.4        & 89.2  & 64.1  & 76.9      \\ 
OAFA \textcolor{red}{(CVPR'24)} \cite{chen2024weakly}                                                  & \multicolumn{1}{c}{}                        & 90.3  & 76.8  & \textbf{73.3}        & 90.3  & \underline{66.0}    & 79.4       \\
\textbf{UAVD-Mamba} (ours)                                                  &\multicolumn{1}{c}{}                                             & \textbf{98.6}     & \textbf{83.9}     & \underline{69.8}           & \textbf{96.9}     &  \textbf{66.1}     & \textbf{83.0}            \\ 
\bottomrule 
\end{tabular}
\end{table*}

\section{Experiments}

\subsection{Experimental Setup}
\textbf{Dataset and Metrics.} We conducted experiments on the DroneVehicle dataset\cite{sun2022drone}, which contains 28,439 visible-infrared image pairs and 953,087 annotated bounding boxes in five categories, including car, truck, freight car, bus, and van. The dataset is divided into 17,990 training sample pairs, 1,469 validation sample pairs, and 8,980 test sample pairs. We use the labels of target objects from modality images with more annotations as the ground truth. Following previous studies\cite{wang2024multi,wang2023directional}, we report the mean average precision (mAP) with an intersection over union (IoU) threshold of 0.5 for evaluation.

\textbf{Implementation Details.} The experiments are carried out on a single NVIDIA RTX 4090 GPU with 24 GB of memory. We implement our algorithm with the PyTorch toolbox and the SGD optimizer with a momentum of 0.937 and a weight decay of 0.0005. The initial learning rate is set to 0.01 and is eventually reduced to 0.0001 by cosine annealing. The batch size is 8. The training epoch is set to 100 epochs. Data augmentation is used to combine four training images into one to simulate different scene compositions and object interactions, and data augmentation is turned off in the last 10 epochs. Before feature extraction, resize the image from 840×712 to 640×640.

\subsection{Results Comparisons}

\textbf{Quantitative comparison.}  The quantitative results are shown in Tab.~\ref{tab:tableTab}. Among the multi-input methods, our UAVD-Mamba benefits from the deformable token specifically designed for Mamba, resulting in a significant improvement in mAP compared to other methods. The mAP value reaches 83.0\%, which is 3.6\% higher than the baseline OAFA\cite{chen2024weakly} method. Additionally, the detection metrics in car, truck, bus and van are the best among all. The detection performance for car is excellent, reaching 98.6\%. Bus also achieved a high detection average precision of 96.9\%. The detection performance for truck is relatively good at 83.9\%, though lower than that for car and bus. The freight car category shows comparatively lower performance of 69.8\%, and van of 66.1\%.

\begin{figure*}[t!]
\vspace{5pt} 
\centering
\setlength{\abovecaptionskip}{-0.1cm}
\includegraphics[width=1\textwidth]{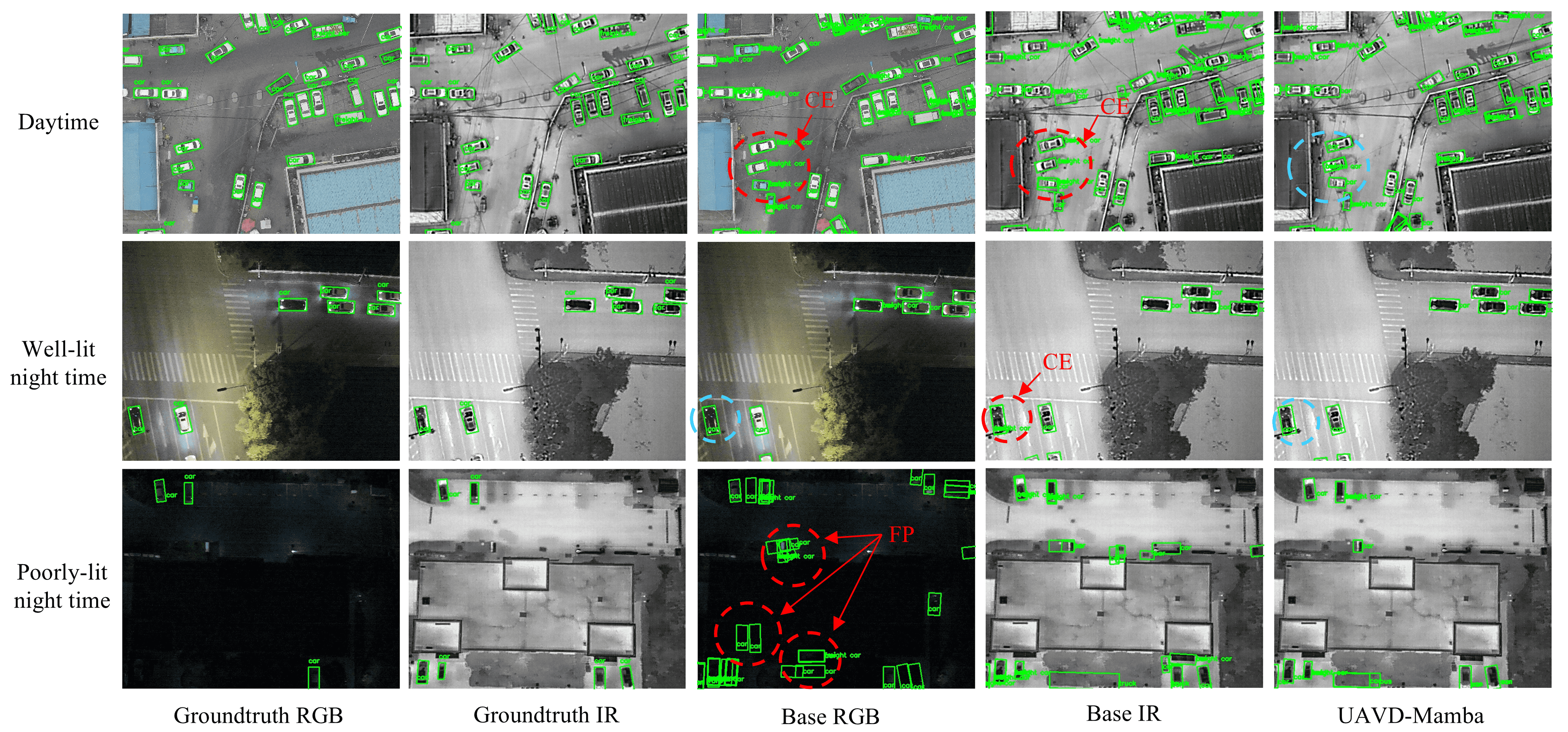}
\caption{Detection results on DroneVehicle dataset. The confidence threshold is set to 0.6. The base model is YOLOv11. We visualize Ground truth in RGB and IR images, and the detection results of Base RGB (3rd column), Base IR (4th column), and our method (5th column). We note that the baseline method, OAFA, is not open-source, and thus we choose Base RGB and Base IR for comparison. Base RGB and Base IR are single-modality methods, and our method is a multimodal fusion method. For simplicity, the detection results of our method, UAVD-Mamba, are visualized in the IR images. There exist several incorrectly detected objects (red dashed circles) in the Base RGB and Base IR, including false positives (FP) and category errors (CE). In contrast, our UAVD-Mamba can correctly detect the objects (blue dashed circles) in those areas, demonstrating our superiority.}
\label{fig:result}
\end{figure*}

\textbf{Qualitative Comparison.}  Our detection model is based on improvements on YOLOv11. Therefore, we use YOLOv11 as the base model and compare our detection results with the base model in RGB and IR modalities. Fig.~\ref{fig:result} shows the visual detection results of our method and the base model. The columns from the first to the last are groundtruth RGB, groundtruth IR, base RGB, base IR, and UAVD-Mmaba. The first row is detection results at daytime, base RGB, and base IR perform poorly in recognizing the correct category. Due to substantial information loss in RGB images at night and the loss of texture in infrared (IR) images, using either RGB or IR images can lead to false positive or category error issues. In contrast, our model fully leverages the rich texture information in multimodal data, exhibiting exceptional detection accuracy, particularly under low-light conditions. This advantage significantly improves the accuracy of recognizing objects with similar shapes while effectively reducing misidentifications in complex low-light environments.

\begin{table}[H]
\centering
\caption{Parameter Size and Computational Loads.} 
\setlength{\tabcolsep}{12pt}
\renewcommand{\arraystretch}{1.5}
\label{tab:Parameter quantity} 
\begin{tabular}{cccc}
\toprule
Method      & mAP (\%)  & Params (M) & GFlops \\ 
\midrule
YOLOv11-RGB & 75.2 & 18.2   & 21.3   \\ 
YOLOv11-IR  & 79.3 & 18.2   & 21.3   \\ 
SLBAF       & 76.6 & 6.3    & 93.3   \\ 
C$^2$Former     & 74.2 & 132.5  & 100.9  \\ 
\textbf{UAVD-Mamba}        & \textbf{83.0} & 39.7   & \underline{38.9}   \\ 
\bottomrule 
\end{tabular}
\end{table}

\begin{table}[H]
\centering
\caption{Inference speed: velocity conversion on DroneVehicle dataset. The base model is YOLOv11. The F denotes FFAR, and D denotes DTMB. UAVD-Mamba-FAST is base+DTMB.} 
\belowrulesep=0pt
\aboverulesep=0pt
\setlength{\tabcolsep}{8pt}
\renewcommand{\arraystretch}{1.5}
\label{tab:Velocity conversion} 
\begin{tabular}{ccc|c}
\toprule
Method         & A6000 (FPS)                   & 4090 (FPS)  & mAP (\%)   \\ 
\midrule
SLBAF          & 63.2                    & 34.0& 76.6  \\ 
OAFA           & 33.1                    &17.8   & 79.4  \\ 
UAVD-Mamba-FAST        & \underline{45.0}  & 24.2   & \underline{81.7}  \\  
\textbf{UAVD-Mamba} & 26.8   & 14.4  & \textbf{83.0}  \\ 
\bottomrule 
\end{tabular}
\end{table}

\subsection{Model Parameter and Inference Speed}
Tab.~\ref{tab:Parameter quantity} shows the model parameter size and floating-point computation load, and our UAVD-Mamba achieves the highest mAP among all object detection methods, with fewer parameters and GFlops, achieving an excellent balance between resource efficiency and detection accuracy. Tab.~\ref{tab:Velocity conversion} shows the inference speed and detection accuracy. To improve the inference speed, we also propose the fast version of UAVD-Mamba, called UAVD-Mamba-FAST, which only includes DTMB, without FFAR and DNM. It achieves 45.0 FPS on the A6000 and 24.2 FPS on the 4090, with a mAP of 81.7\%, outperforming the multimodal SOTA method OAFA\cite{chen2024weakly}. This demonstrates significant potential for the practical application of UAV object detection.

\begin{figure}[H]
    \centering
    \includegraphics[width=1\linewidth]{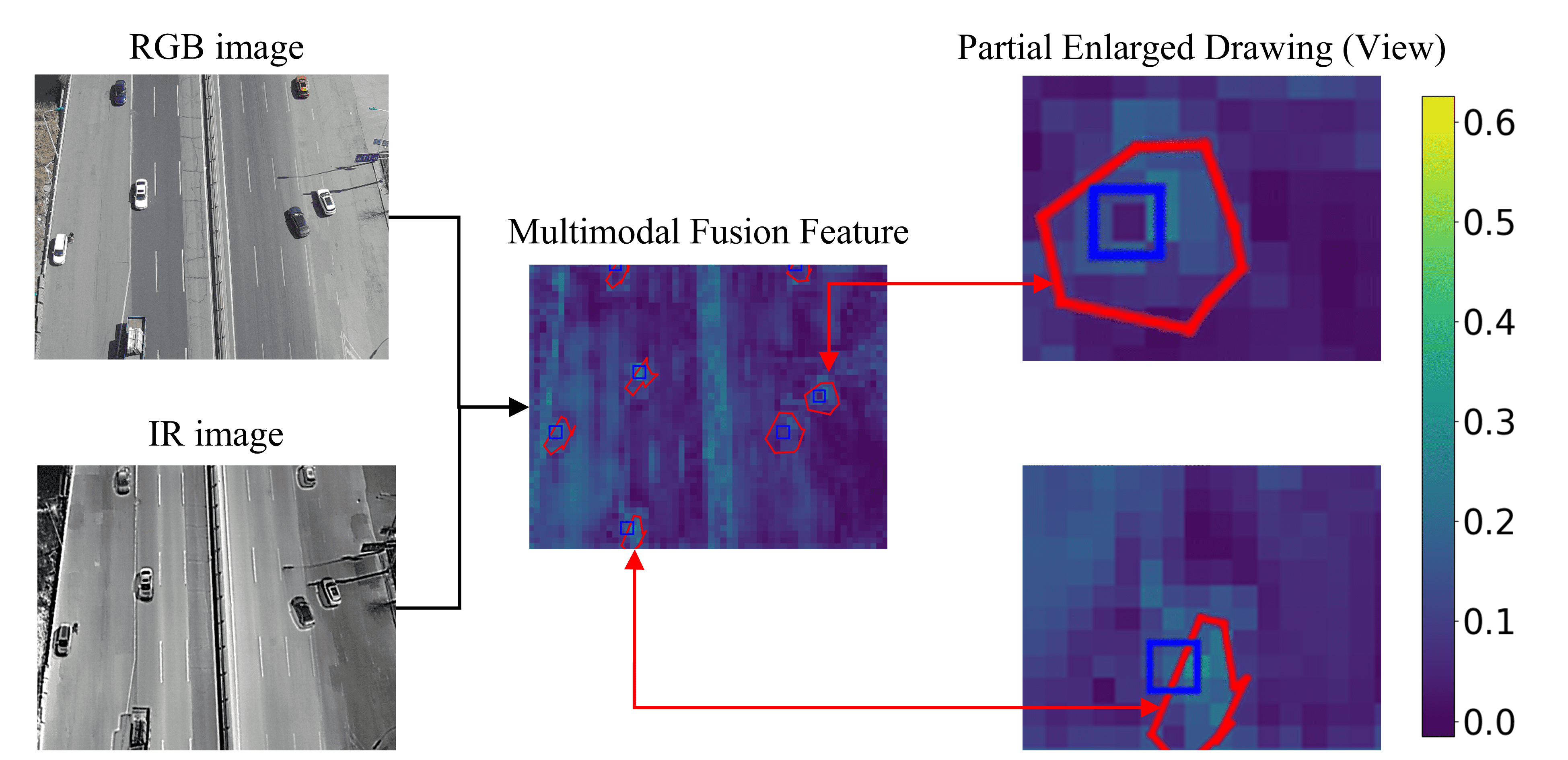}
    \caption{Visualization of the normal patches (blue) and the adaptive patches (red), shown in the partially enlarged drawing (view). For simplicity, we draw the patches near vehicles for demonstration. Normal patches are square-shaped, while our patches have irregular shapes, allowing them to better adapt to targets of varying shapes. In this way, the deformable image tokens generated by adding normal patches and adaptive patches can capture more discriminative features for the Mamba blocks while retaining the information of the normal patches.}
    \label{fig:feature}
\end{figure}

\subsection{Visualization} The adaptive patches (red) and the normal patches (blue) near the vehicles are visualized in Fig.~\ref{fig:feature}, and RGB and IR modalities are used to generate multimodal fusion features. Normal patches have a smaller scope with a square shape, capturing only partial image features. In contrast, adaptive patches can adaptively adjust the shape of the patch and can extract important feature regions. We add the normal patch and the adaptive patch to generate the deformable tokens, which can capture more discriminative features while retaining the information of the normal patch.

\begin{figure*}
\vspace{5pt} 
\centering
\setlength{\abovecaptionskip}{-0.1cm}
\includegraphics[width=1\textwidth]{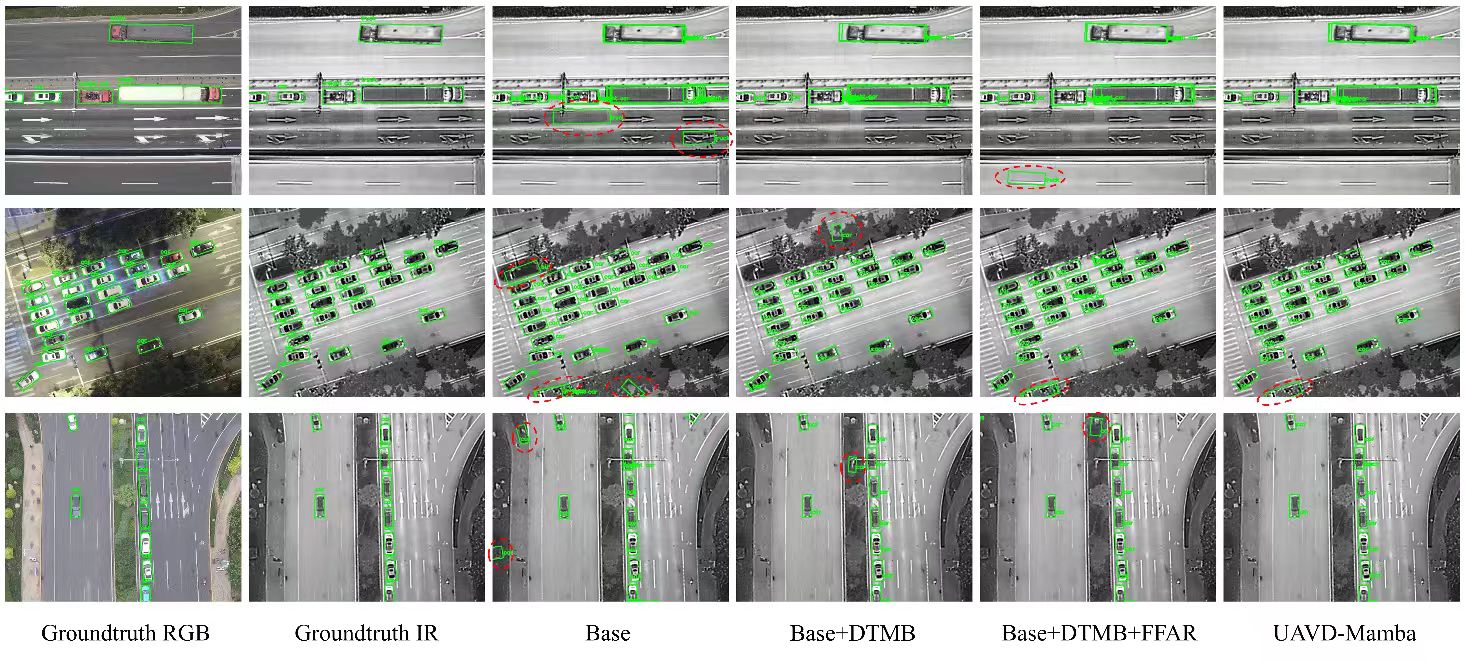}
\caption{Ablation experiment detection results on DroneVehicle dataset. The confidence threshold is set to 0.6. The results of our complete method are in the last column, with the fewest false positive samples, proving the effectiveness of our approach.}
\label{fig:Ablation}
\end{figure*}

\subsection{Ablation Experiment}
Tab.~\ref{tab:Ablation Experiment} presents the ablation results, highlighting the effectiveness of the DTMB, FFAR, and DNM modules on the performance of our UAVD-Mamba. The base model for comparison is YOLOv11. Initially, we evaluate the standalone performance of the DTMB module. Adding DTMB to YOLOv11 leads to an improvement in mAP from 79.6\% to 81.7\% (+2.1\%). Next, incorporating the FFAR module into the base+DTMB configuration further boosts the mAP by 2.7\% compared to the base model. Finally, optimizing the YOLO detection neck with the DNM module on top of base+DTMB+FFAR results in a 3.4\% improvement. Notably, the DTMB module contributes the most to the performance gains. These results demonstrate the effectiveness of the DTMB, FFAR, and DNM modules in enhancing the accuracy of our UAVD-Mamba model.

\begin{table}[htpb]
\centering
\caption{Ablation study on DroneVehicle dataset. The base model is YOLOv11. F denotes FFAR, D denotes DTMB.} 
\belowrulesep=0pt
\aboverulesep=0pt
\setlength{\tabcolsep}{8pt}
\renewcommand{\arraystretch}{1.5}
\label{tab:Ablation Experiment} 
\begin{tabular}{cccc|c}
\toprule
Method   & DTMB & FFAR & DNM & mAP (\%)  \\ 
\midrule
Base &      &      &      & 79.6 \\ 
Base+D & \textcolor{red}{\checkmark}    &      &      & 81.7 (+2.1\%) \\ 
Base+D+F         & \textcolor{red}{\checkmark}    & \textcolor{red}{\checkmark}    &      & 82.4 (+2.7\%) \\ 
\textbf{UAVD-Mamba}    & \textcolor{red}{\checkmark}    & \textcolor{red}{\checkmark}    & \textcolor{red}{\checkmark}    & \textbf{83.0 (+3.4\%)} \\ 
\bottomrule 
\end{tabular}
\end{table}

As shown in Fig.~\ref{fig:Ablation}, UAVD-Mamba excels in detecting occluded and small targets while effectively reducing false detections of similar objects such as trees, roads, and lane markings. Its performance improvements stem from several key optimizations: DTMB, combined with deformable convolutions and multiscale stacking, enhances the detection of occluded and small targets; cross-enhanced spatial attention and cross-channel attention improve feature differentiation, enabling more accurate target recognition while minimizing background interference. Additionally, the independent processing of RGB and infrared data, integrated with Mamba Block for feature fusion, maximizes the utilization of multimodal information, allowing UAVD-Mamba to maintain high detection accuracy even in complex environments.

\subsection{Limitation}
\begin{figure}[]
    \centering
    \setlength{\abovecaptionskip}{-0.1cm}
    \includegraphics[width=1\linewidth]{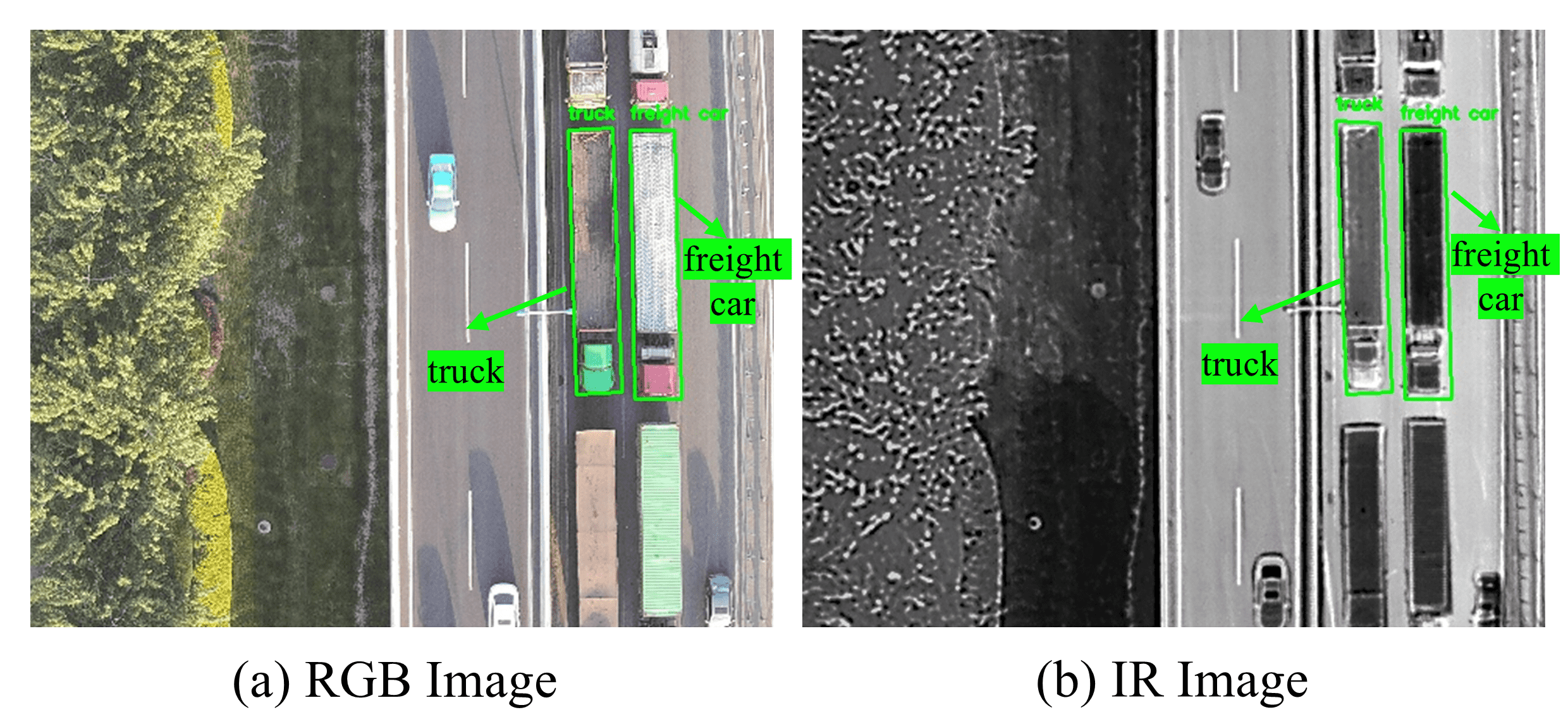}
    \caption{Illustration of freight cars and trucks in RGB and IR images. There are significant similarities between freight cars and trucks, making it difficult to distinguish between the two categories.}
    \label{fig:limitation}
\end{figure}

We observed that the detection accuracy of freight cars in UAVD-Mamba is lower than that of OAFA. As shown in Fig.~\ref{fig:limitation}, due to the similar shapes between freight cars and trucks, it is difficult to clearly distinguish between the two using IR images alone. Although RGB images offer texture information, distinguishing between the two categories remains challenging for our method when the texture details are insufficient, even for human annotators. Moreover, the limited number of freight car labels also degrades accuracy. In future work, we will focus on utilizing the texture information in RGB images and few-shot multimodal fusion to improve the detection accuracy of freight cars.

\section{Conclusion}
In this paper, we propose UAVD-Mamba, a multimodal UAV object detection framework based on Mamba architectures. We generate adaptive deformable tokens for Mamba Blocks to enhance the feature extraction of objects with irregular shapes. By designing separate Deformable Token Mamba Blocks (DTMB) for RGB and infrared (IR) modalities, we can improve the multimodal feature complementarity. Additionally, incorporating a multiscale detection neck for mamba and modifications to YOLOv11's SPPF and C3K2 components further strengthen feature processing, enhancing object detection performance across diverse scales and modalities. Our method can achieve higher accuracy with fewer parameters while reducing data redundancy. Future work focuses on few-shot learning for multimodal UAV detection.


\bibliographystyle{IEEEtran} 
\bibliography{ref} 
\vspace{12pt}

\end{document}